\documentclass[runningheads]{llncs}
\usepackage{wrapfig, booktabs}
\usepackage{graphicx}
\usepackage{subfigure}
\usepackage{amsmath,amssymb}
\usepackage{times}
\usepackage{epsfig}
\usepackage{graphicx}
\usepackage{amsmath}
\usepackage{amssymb}
\usepackage{dirtytalk}
\usepackage{xcolor}
\usepackage{cite}
\usepackage{graphicx}
\usepackage{subfigure}
\usepackage{amsmath,amssymb}

\begin{document}
\title{Unsupervised Shape Normality Metric for Severity Quantification}
%
%
\author{Wenzheng Tao\inst{1} \and
Riddhish Bhalodia\inst{1,2} \and
Erin Anstadt\inst{3}\and
Ladislav Kavan\inst{1}\and
Ross T. Whitaker\inst{1,2}\and
Jesse A. Goldstein\inst{4}}
\institute{School of Computing, University of Utah \and
Scientific Computing and Imaging Institute, University of Utah \and
Department of Plastic Surgery, University of Pittsburgh \and
Children's Hospital of Pittsburgh}
%
\maketitle              
\begin{abstract}
This work describes an unsupervised method to objectively quantify the abnormality of general anatomical shapes. The severity of an anatomical deformity often serves as a determinant in the clinical management of patients. However, experiential bias and distinctive random residuals among specialist individuals bring variability in diagnosis and patient management decisions, irrespective of the objective deformity degree. Therefore, supervised methods are prone to be misled given insufficient labeling of pathological samples that inevitably preserve human bias and inconsistency.  Furthermore, subjects demonstrating a specific pathology are naturally rare relative to the normal population. To avoid relying on sufficient pathological samples by fully utilizing the power of normal samples, we propose the shape normality metric (SNM), which requires learning only from normal samples and zero knowledge about the pathology. We represent shapes by landmarks automatically inferred from the data and model the normal group by a multivariate Gaussian distribution. Extensive experiments on different anatomical datasets, including skulls, femurs, scapulae, and humeri, demonstrate that SNM can provide an effective normality measurement, which can significantly detect and indicate pathology. Therefore, SNM offers promising value in a variety of clinical applications.

\end{abstract}

\section{Introduction}
\label{sec:intro}

Many medical conditions are indicated by pathological shapes, such as abnormal skull shape (craniosynostosis), hip deformities (femoroacetabular impingement), and shoulder dislocation  (scapula and humerus injury). Current medical imaging methods aim to identify abnormal anatomical configurations and guide physicians toward diagnosis and treatment plans; severe cases may warrant surgical intervention, while observation might suffice for mild deformities. But how can one objectively quantify the severity of a pathological three-dimensional (3d) shape deformity? There is no easy answer, because "normal" anatomy represents a spectrum, comprised of a variety of different shapes and sizes due to genetic and environmental factors, that does not necessarily result in pathology or compromised function.

Current clinical practice frequently relies on severity assessment by medical professionals. However, despite the best efforts of clinicians and their rigorous medical training, there is a concern for bias in individual clinicians\cite{sica2006bias},due to variables such as specific training sites or years of practice. For example, there is a wide disparity of opinion regarding the diagnosis of mild nonsyndromic metopic synostosis. \cite{yee2015classification}. 
The concern of bias applies also to supervised machine learning approaches \cite{bhalodia2019severity}, because the bias in the training data may translate to bias in the model predictions. Some types of biases are well studied (e.g., discrimination due to gender or ethnicity), but the biases in evaluating shape pathologies are more subtle and are poorly understood. Limited number of pathological samples, especially with relatively uncommon diseases, such as craniosynostosis, poses a problem of data skewness in supervised analysis. 

To circumvent these issues, in this paper we explore an \textit{unsupervised} approach to quantify the severity of shape pathologies. By unsupervised, we mean that our method relies solely on examples of shapes of normal anatomical variations. They are provided as point correspondences that describe normal shape variations in a population (as in point distribution models \cite{cootes2004statistical,cates2007shape,oguz2016entropy}). We don't use any example shapes of a given pathology (such as metopic craniosynostosis), which means that our method is fully general and agnostic to any specific type of pathologies. Because it merely quantifies the distance from normal anatomical variations, we call it ``Shape Normality Metric'' (SNM). 

We found that even without being informed by examples of specific pathologies, SNM is very useful in predicting severity of shape deformity, as we validated by retroactively comparing the SNM results with evaluations by human experts. However, compared with supervised methods, the SNM method has distinct advantages: no pathological training data required, thus immune to any biases present in such data. We foresee a practical utility of SNM in clinical decision making -- not as a method to automatically issue a diagnosis, but rather by providing a new tool to the clinicians, akin to a more sophisticated ''measuring tape'' (one which measures statistical shape differences). 





\section{Related work}
\label{sec:literature}

Unsupervised severity quantification makes heavy use of statistical shape modeling \cite{thompson1942growth}. It involves representing a set of 2D/3D shapes such that it captures the population statistics, followed by application-specific analysis. A popular way of shape representation is via correspondence based methods/particle distribution models \cite{RTW:Gre91}. Correspondences are geometrically consistent points placed manually or automatically on the set of shapes, simple to use and convenient to parametrize shape for subsequent statistics. Recent years have seen a rise of methods that automatically places a dense set of correspondences on a population of anatomical shapes \cite{styner2006spharm, davies2002MDL, cates2007shape}. The usual methodology of analyzing dense correspondence representation is to express them in a low-dimensional space via principal component analysis (PCA) \cite{cates2008particle} and use that as the shape descriptor. 
These methods have proved useful in many fronts of medical imaging such as orthopedics \cite{harris2013cam} and cardiology \cite{cates2013afib}. They have also been used in parametrizing skull shape \cite{RTW:Dat2009}, serving as a platform for our proposed modeling on pediatric skulls. 

Also related is the field of unsupervised anomaly detection. These methods usually learn the distribution of normal samples and detect the anomaly by measuring its distance/dissimilarity. One of the most common ways to detect anomalies from a given distribution is via the use of Mahalanobis distance. A robust variant of the Mahalanobis distance based on the Minimum Covariance Determinant has been proposed, tackling the sensitivity to outliers in observed normal samples \cite{leys2018detecting}. Another branch adopts deep learning related techniques while avoiding its preconditions on a large number of pathological samples. AnoGAN is proposed to identify anomalous images, composed of a Generative Adversarial Networks together with a novel anomaly scoring scheme \cite{schlegl2017unsupervised}. Further, Auto-encoder based methods are utilized for the detection of lesion regions by learning data distribution of healthy brain MRIs \cite{chen2018unsupervised}. Recently, Variational Auto-Encoders have been used to check and localize suspicious parts within images \cite{zimmerer2019unsupervised}.


There have been efforts to quantify the severity of metopic craniosynostosis by parametrizing cranial shapes. Diagnosis is particularly challenging as the metopic suture normally closes before 1 year of age \cite{weinzweig2003metopic} which can limit the methods for detecting suture fusion (used for other forms of craniosynostosis) \cite{eley2014black, mendoza2014personalized}. Common approaches thus far involve defining a simple geometric measure on the skull and correlate that with the craniosynostosis severity, the most popular of these being the \emph{interfrontal angle (IFA)} \cite{kellogg2012interfrontal}, which captures the degree of trigonocephaly (triangular frontal head shape). IFA can be calculated from the CT scans and has been effective in distinguishing normal and metopic head shapes \cite{wood2016name, kellogg2012interfrontal}, though its utility is limited in predicting the severity among abnormal shapes. Recent work used expert supervision with an expressive shape descriptor to measure the craniosynostosis severity \cite{bhalodia2019severity}, however, concerns exist about biases in the export ratings.

\section{Materials and methods}
\label{sec:methods}

By denoting the $i$-th 3D shape as $S_i^{(b)} \in \mathcal{S} \subset R^3$, where $b \in \{N,P\}$ indicates whether the shape is normal $(N)$ or pathological $(P)$, we have the shape observations $\{S_i^{(b)}\} $.

Shapes are continuous surfaces and thus inconvenient for calculation. With Shapeworks \cite{cates2017shapeworks}, we obtain the particle-based representation. Specifically, each shape $S_i^{(b)}$ is described by a cloud of $\frac{p}{3}$ ordered particles serving as correspondences across different shape samples. This particle cloud is denoted as $C_i^{(b)} \in R^{p \times 1}$, a flattened vector of these $\frac{p}{3}$ correspondences' coordinates $\{(x_k, y_k, z_k)_i\} \in R^{3 \times \frac{p}{3}}$. 

\begin{figure}
\centering
\minipage{1\textwidth}
  \includegraphics[width=\linewidth]{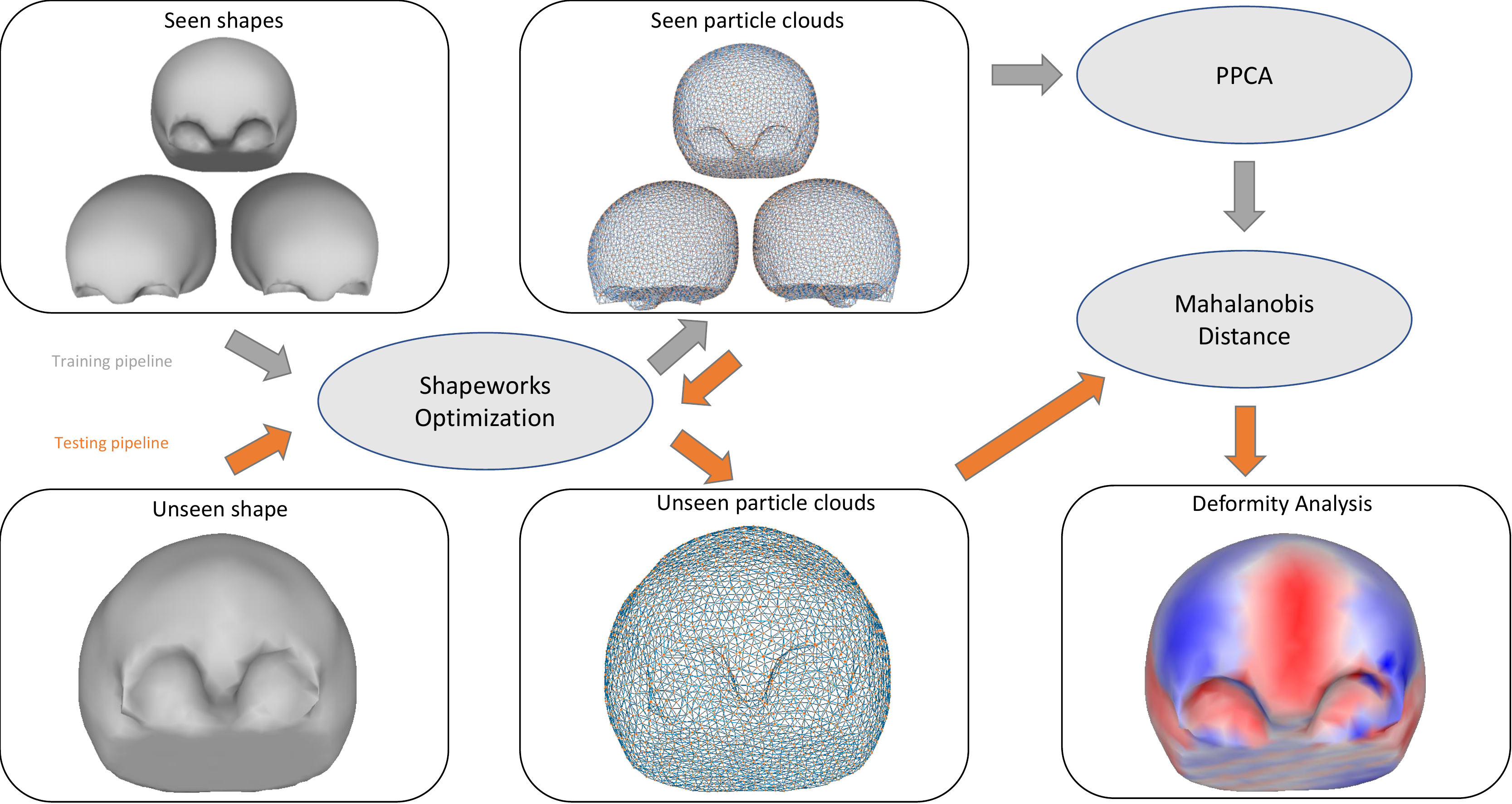}
\endminipage\hfill
\caption{Flow chart of SNM with training and testing pipelines.}
\label{fig:pipelines_snm}
\end{figure}

As shown in Figure \ref{fig:pipelines_snm}, there is 1) training, and 2) inference (testing) pipeline. In training, we used Shapeworks to obtain correspondences of present normal shapes which will be frozen during inference. Meanwhile, Shapeworks estimates density function of these correspondences $\{\Pr [C_i^{(N)}] \}$. Next for each new shape $S_{new}^{(b)}$ where $b \in \{N,P\}$ is unknown, we run Shapeworks on $S_{new}^{(b)}$ jointly with previously frozen normal correspondences $\{C_i^{(N)}\}$. This way we obtain the correspondences $C_{new}^{(b)}$ for shape $S_{new}^{(b)}$, as the optimal landmark points under the density function of $\{C_i^{(N)}\}$, which both faithfully encode the shape information of $S_{new}^{(b)}$ and are consistent with $\{C_i^{(N)}\}$.

With the shapes represented in the form of coordinate vectors, we can apply probabilistic methods to model the shape variations. To consider both seen and unseen variations with limited normal samples for training, we adopt the Probabilistic Principal Component Analysis (PPCA) \cite{bishop2006pattern}.

\subsection{Probabilistic Principle Component Analysis}

Assume we have $n$ normal observations $C^{(N)} \in R^{p \times n}$, where the $i$-th observation is a $p$-dimensional feature vector $C_i^{(N)} \in R^{p \times 1}$. In the generative view of PPCA, we assume that the principle normal shape variations originate from a low-dimensional vector in a compact latent space $L_i \in \mathcal{L} \subseteq R^{d \times 1}$, with $d \leq (n-2) \leq p$. Specifically,
\begin{equation}
    C_i^{(N)} = WL_i + \mu + \epsilon \sim \mathcal{N}(\mu, WW^T + \sigma^2I)
\end{equation}
where $L_i$ obeys standard multivariate Gaussian distribution $ \mathcal{N}({0}, {I})$, $W \in R^{p \times d}$ is the weighting matrix controlling the linear transformation, $\mu = \mathop{\mathbb{E}} [C_i^{(N)}]$ is the expectation of normal shape's correspondences, and $\epsilon \sim \mathcal{N}({0}, { \sigma^2I})$ is the $p$-dimensional noise.







With Maximum Likelihood Estimation, we have the latent parameters estimated from normal observations:
\begin{equation}
    \mu_{ML} = \frac{\sum_{i=1}^n C_i^{(N)}}{n}, W_{ML} = U_d( \Lambda_d - \sigma_{ML}^2 I)^{\frac{1}{2}} O, \sigma^2_{ML} = \frac{\sum_{i=d+1}^{n-1} \lambda_i}{n-1 - d}
\end{equation}
$\{\lambda_i\}$ are the descending eigen-values of the empirical co-variance matrix $\frac{\Bar{C}^{(N)}{\Bar{C}^{(N)T}}}{n-1}$, $\Bar{C}^{(N)} = C^{(N)}-\mu_{ML}$, $\Lambda_d \in R^{d \times d} $ is the diagonal matrix with $d$ largest $\{\lambda_i\}$, $U_d \in R^{p \times d}$ has the corresponding eigen-vectors as its columns, and $O \in R^{d \times d}$ is an arbitrary orthogonal matrix. As $O$ does not affect $(WW^T + \sigma^2I)$ we simply set $O$ to the identity.

\subsection{Shape Normality Metric}

\subsubsection{Overall Mahalanobis Distance}

To measure the abnormality of $C_{new} \in R^{p \times 1}$, we calculate its Mahalanobis distance as the Shape Normality Metric (SNM):
\begin{equation}
    SNM(C_{new}) = \sqrt{(C_{new} - \mu_{ML})^T(W_{ML}W_{ML}^T + \sigma_{ML}^2I)^{-1}(C_{new} - \mu_{ML})}
    \label{eq:md_whole}
\end{equation}
The distance in Equation \ref{eq:md_whole} consists of 2 components. First is the Mahalanobis distance in the $d$-dimensional latent space $\mathcal{L}$, measuring deformity in normal variations:
\begin{equation}
    SNM_d(C_{new}) =  \sqrt{(C_{new} - \mu_{ML})^TU_d(\Lambda_d + \sigma_{ML}^2I)^{-1}U_d^T(C_{new} - \mu_{ML})}
    \label{eq:md_column}
\end{equation}
The other is the normalized distance to $\mathcal{L}$ in the $(p-d)$-dimensional null space, measuring deformity in abnormal variations undetected by the training normal shapes:
\begin{equation}
    SNM_{\epsilon}(C_{new}) =  \sqrt{\sigma_{ML}^{-2}||(C_{new} - \mu_{ML})^TU_{\epsilon}||_2^2}
    \label{eq:md_null}
\end{equation}
where $U_{\epsilon} \in R^{p \times (p-d)}$ has columns as orthonormal null vectors to $U_d$. It generalizes SNM for unseen deformity, with its superiority corroborated in our experiments.

\subsubsection{Point-wise Mahalanobis Distance.}

Equation \ref{eq:md_whole} can also be regarded as a $L2$-norm, of the new shape's whitened deviations from mean:
\begin{equation}
    (W_{ML}W_{ML}^T + \sigma_{ML}^2I)^{-\frac{1}{2}}(C_{new} - \mu_{ML})
\end{equation}
With whitened deviations in $k$-th correspondence coordinates expressed as $(\tilde{x}_k, \tilde{y}_k, \tilde{z}_k)$, we use the whitened deviation's $L2$-norm to measure the shape's local deformity:
\begin{equation}
    ||(\tilde{x}_k, \tilde{y}_k, \tilde{z}_k)||_2
\label{eq:normalize_point}
\end{equation}
Equation \ref{eq:normalize_point} normalize the deviation on different correspondences to lie in the same variance scale so that they are comparable.

\section{Results}
\label{sec:results}

In this section, we present experimental results on various medical conditions demonstrating that SNM provides accurate deformity scores, which are strongly indicative of general pathology deformations. For Metopic Craniosynostosis, SNM outperforms deformity ratings issued by human experts and is comparable with their aggregation, according to Area Under Curve (AUC) \cite{hanley1982meaning} with the known diagnosis as ground truth. 


\subsection{Metopic Craniosynostosis}

Metopic Craniosynostosis is characterized by skull deformation due to the premature fusion of the metopic suture. Affected individuals typically exhibit trigonocephaly with the classic triad of a narrow "keel-shaped" forehead, biparietal widening, and hypotelorism.

\subsubsection{Metopic Craniosynostosis dataset}

We aggregated 124 head CT scans of patients between 6-16 months old. 30 (24\%) of these patients are diagnosed with Metopic Craniosynostosis; while the remaining 94 (76\%) are trauma patients with no known intracranial or calvarial abnormality (normal patients). All 124 scans were segmented as outer skull surfaces. We used 74 randomly selected normal skull shapes for training, and left 20 normal shapes, as well as all 30 pathological shapes, for testing.  

To collect the ground truth deformity degrees of the 50 testing skulls, we displayed their 3d segmentation on our website and asked physicians to estimate their deformity on a 5 point Likert scale. In total, 31 physicians participated in the survey, among them 14 rated all of the 50 skulls and the other 17 rated 20 skulls randomly. We modeled the expert ratings by Latent trait theory \cite{uebersax1993latent, muthen1998statistical} which accounts for potential raters' bias and inconsistency and estimated the continuous latent severity of the 50 skulls using Maximum Likelihood Estimation (MLE). The AUC of diagnosis and ratings from the 14 physicians as individuals has mean of 0.8860 and standard deviation of 0.03426, with the highest one being 0.9458. AUC of diagnosis and MLE severity is 0.9850, indicating that aggregation of physicians estimations is more accurate than individuals.


\subsubsection{Severity predictions}

We trained SNM on 74 normal skulls and predicted deformity scores for the 50 testing skulls. To unify the skulls as inputs for SNM, we cropped the skulls above the plane defined by the nasion and two porion points. Next, they went through 2 pipelines in Figure \ref{fig:pipelines_snm}. In the training pipeline, we ran Shapeworks optimization on the 74 normal skulls jointly, obtained their correspondences of 2048 particles, calculated their mean and covariance matrix using PPCA. PPCA requires specification of the latent subspace dimension and by convention we set it as the dimension where it explains 95\% of the variance. In the testing pipeline, for every testing skull, we fed it into Shapeworks together with the frozen normal correspondences, got its correspondences and applied deformity analysis by calculating Mahalanobis Distance.

\begin{figure}
\centering
\minipage{1\textwidth}
  \includegraphics[width=\linewidth]{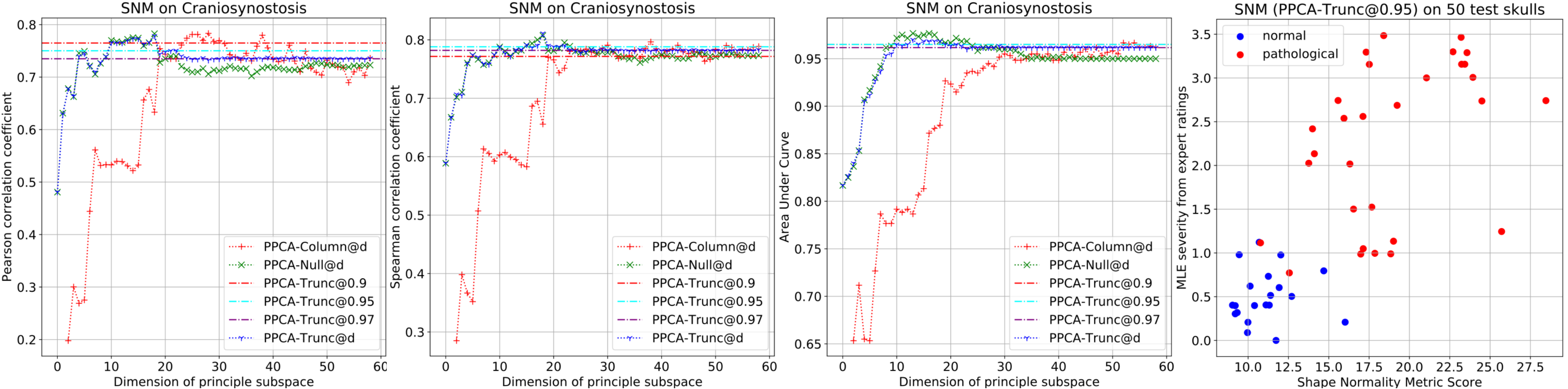}
\endminipage\hfill
\caption{Evaluation of Shape Normality Metric on Craniosynostosis. [PPCA-Column@d] is using the Mahalanobis distance in the $d$-dimensional latent subspace $\mathcal{L}$ as Equation \ref{eq:md_column}. [PPCA-Null@d] is using the Mahalanobis distance in the null space as Equation \ref{eq:md_null}. [PPCA-Trunc@$\alpha$] is using the Mahalanobis distance in the whole space as Equation \ref{eq:md_whole}, where $\alpha$ is either principle subspace dimension or explained variance ratio. Our SNM is [PPCA-Trunc@0.95]}
\label{fig:evaluation_chart1}
\end{figure}

\subsubsection{Correlation analysis}

The agreement between SNM and the continuous latent severity is evaluated by Pearson Correlation Coefficient \cite{benesty2009pearson} and Spearman Correlation Coefficient \cite{kokoska2000crc}. The agreement between SNM and the diagnosis labels is evaluated by AUC. 

The results are shown in Figure \ref{fig:evaluation_chart1}. Our SNM is [PPCA-Trunc@0.95], which achieves the Pearson Correlation Coefficient of 0.7501, Spearman Correlation Coefficient of 0.7881 and AUC of 0.965, showing strong agreement with the ground truth. Also, AUC is even higher than the highest of our 14 physicians, which is encouragingly showing that our SNM is a better indicator in the case of Craniosynostosis compared with deformity estimations by individual physicians.

\begin{figure}
\centering
\minipage{1\textwidth}
  \includegraphics[width=\linewidth]{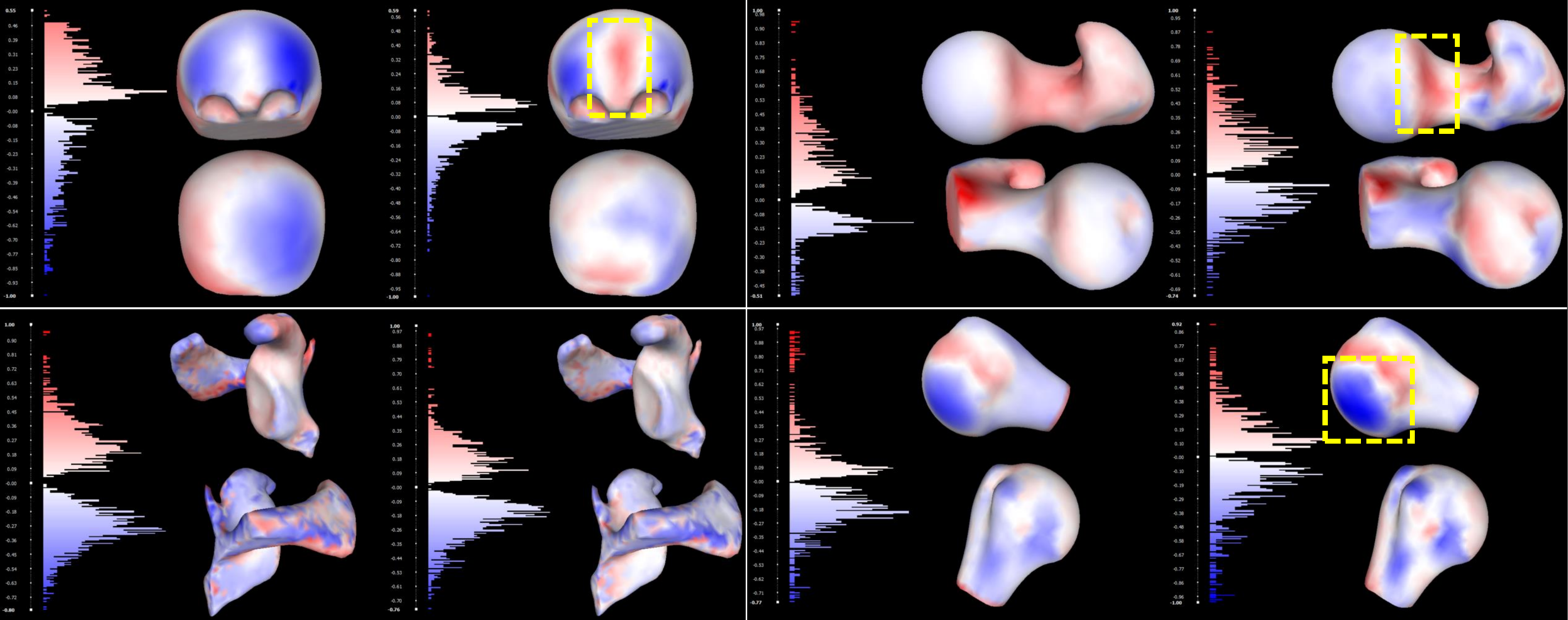}
  \label{fig:deviation}
\endminipage\hfill
\caption{Unified point-wise signed distance on the same pathological skull, femur, scapula and humerus: original  (left) and whitened by SNM (PPCA-Trunc@0.95) using Equation \ref{eq:normalize_point} (right). Whitened is more symmetric and accentuated the principal pathological pattern (in yellow frame).}
\label{fig:snm_deviations}
\end{figure}

We also compare the proposed SNM metric with several baselines. First, [PPCA-Null@0] is essentially calculating the Mahalanobis distance with an isotropic covariance matrix and it has relatively poor performance. This indicates that it is disadvantageous to treat all the deformation components with the same variance. It can also be shown in Figure \ref{fig:snm_deviations}, where different point-wise deviations of the same testing skull are compared using MeshLab \cite{cignoni2008meshlab}. In the one whitened by SNM, point-wise signed distance is more symmetric and similar to root of Chi-squared distribution, and abnormal variations are accentuated on the forehead (Metopic Craniosynostosis pattern). Figure \ref{fig:evaluation_chart1} also shows that SNM is not very sensitive to hyper-parameter settings, as 3 different explained variance ratios generate similar performance which is close to the best presented.


\begin{figure}
\centering
\minipage{1\textwidth}
  \includegraphics[width=\linewidth]{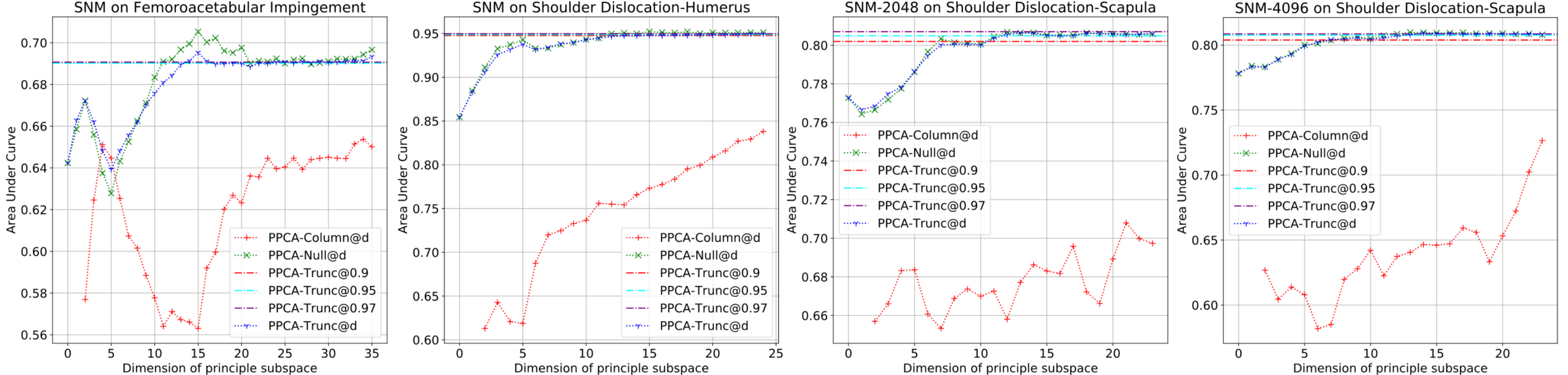}
\endminipage\hfill
\caption{Evaluation of Shape Normality Metric on Femoroacetabular impingement, Shoulder Dislocation-Humerus and Scapula. Notation is the same as Figure \ref{fig:evaluation_chart1}}
\label{fig:evaluation_chart2}
\end{figure}

\subsection{Cam-FAI and Shoulder Dislocation}

Cam-type femoroacetabular impingement can cause hip pain and is characterized by Pincer morphology with acetabular overcoverage \cite{hack2010prevalence}. We collected 37 femur bones with Cam-FAI and 59 normal ones. 

The type of shoulder dislocation we studied is characterized by aberrant positioning of the superior portion of the humerus away from its usual location in the glenohumeral joint (the ball and socket joint of the shoulder). It can result in bone deformation located primarily at the joint. We collected 53 humerus scans and 54 scapula scans from patients diagnosed with shoulder dislocation, as well as 41 normal humerus and scapula scans. These respectively formed our humerus and scapula datasets.

For the 3 datasets above, we segmented the scans to extract the outer bone surfaces and cropped them uniformly. We evaluated SNM on each of the 3 datasets using 3-repeat 3-fold cross-validation. Specifically, we randomly shuffled the normal samples data, divided it into 3 folds, trained SNM on each pair of 2 folds, predicted severity on the remaining fold along with pathological samples and calculated the AUC with the predictions and diagnosis. This entire process is repeated 3 times. For each dataset, we report the average of these 9 AUC scores.

As is shown in Figure \ref{fig:evaluation_chart2}, on all the 3 datasets, [PPCA-Column@d] methods are much worse than the others, indicating that the informative variations mainly lie in the null space which is not captured by the training samples. This indicates it is an advantage of SNM to calculate Mahalanobis distance in the entire space. The results also show that SNM is not sensitive to the hyper-parameters. For the explained variance ratio, all 3 different explained variance ratios are generating nearly the best scores. For the number of particles in Shapeworks, we used 1024 particles for Femur and Humerus, and we did 2 parallel experiments using 2048 and 4096 particles for Scapula. All the numbers are providing visually adequate and smooth coverage of the shape. As shown in Figure \ref{fig:evaluation_chart2}, SNM is not sensitive to the number of particles as well. And the whitened deviations are shown in Figure \ref{fig:snm_deviations}, where the pathological variations are highlighted after whitening. While it is hard to find a predominant pattern for the scapula, maybe due to its complicated structure and deformation patterns.

\section{Conclusions}
\label{sec:conclusions}

We proposed Shape Normality Metric (SNM) as an objective and unsupervised method to measure shape deformity. SNM employs particle-based shape representation and applies Probabilistic Principle Component Analysis to calculate Mahalanobis distance in the whole space, properly accounting for both seen and unseen variations. SNM does not require any data or information about pathological samples and is thus unaffected by pathological population size limitations and/or bias in ratings issued by human experts.  Extensive experiments on real-world datasets validated that SNM provides an accurate and reproducible metric for a wide range of shape deformity quantification. Therefore, SNM has the potential to be applied to a variety of clinical scenarios, with particular promise in objectifying communication on shape deformations and planning interventions. 

\textbf{Acknowledgements : } This work was supported by the National Institutes of Health under grant num- bers NIBIB-U24EB029011, NIAMS-R01AR076120, NHLBI-R01HL135568, and NIGMS-P41GM103545. The content is solely the responsibility of the authors and does not necessarily represent the official views of the National Institutes of Health. Authors would like to thank Andrew Anderson, PhD, Heath B. Henninger, PhD, and Matthijs Jacxsens, MD for providing Scapula and femur shapes.

\bibliographystyle{splncs04}
\bibliography{cranio}

\begin{thebibliography}{10}
\providecommand{\url}[1]{\texttt{#1}}
\providecommand{\urlprefix}{URL }
\providecommand{\doi}[1]{https://doi.org/#1}

\bibitem{benesty2009pearson}
Benesty, J., Chen, J., Huang, Y., Cohen, I.: Pearson correlation coefficient.
  In: Noise reduction in speech processing, pp.~1--4. Springer (2009)

\bibitem{bhalodia2019severity}
Bhalodia, R., Dvoracek, L.A., Ayyash, A.M., Kavan, L., Whitaker, R., Goldstein,
  J.A.: Quantifying the severity of metopic craniosynostosis: A pilot study
  application of machine learning in craniofacial surgery. Journal of
  Craniofacial Surgery  (2020)

\bibitem{bishop2006pattern}
Bishop, C.M.: Pattern recognition and machine learning. springer (2006)

\bibitem{cates2017shapeworks}
Cates, J., Elhabian, S., Whitaker, R.: Shapeworks: Particle-based shape
  correspondence and visualization software. In: Statistical Shape and
  Deformation Analysis, pp. 257--298. Elsevier (2017)

\bibitem{cates2008particle}
Cates, J., Fletcher, P.T., Styner, M., Hazlett, H.C., Whitaker, R.:
  Particle-based shape analysis of multi-object complexes. In: International
  Conference on Medical Image Computing and Computer-Assisted Intervention. pp.
  477--485. Springer (2008)

\bibitem{cates2007shape}
Cates, J., Fletcher, P.T., Styner, M., Shenton, M., Whitaker, R.: Shape
  modeling and analysis with entropy-based particle systems. In: Biennial
  International Conference on Information Processing in Medical Imaging. pp.
  333--345. Springer (2007)

\bibitem{chen2018unsupervised}
Chen, X., Konukoglu, E.: Unsupervised detection of lesions in brain mri using
  constrained adversarial auto-encoders. arXiv preprint arXiv:1806.04972
  (2018)

\bibitem{cignoni2008meshlab}
Cignoni, P., Callieri, M., Corsini, M., Dellepiane, M., Ganovelli, F.,
  Ranzuglia, G.: Meshlab: an open-source mesh processing tool. In: Eurographics
  Italian chapter conference. vol.~2008, pp. 129--136 (2008)

\bibitem{cootes2004statistical}
Cootes, T.F., Taylor, C.J., et~al.: Statistical models of appearance for
  computer vision (2004)

\bibitem{RTW:Dat2009}
Datar, M., Cates, J., Fletcher, P.T., Gouttard, S., Gerig, G., Whitaker, R.:
  Particle based shape regression of open surfaces with applications to
  developmental neuroimaging. In: International Conference on Medical Image
  Computing and Computer-Assisted Intervention. pp. 167--174. Springer Berlin
  Heidelberg (2009)

\bibitem{davies2002MDL}
Davies, R.H., Twining, C.J., Cootes, T.F., Waterton, J.C., Taylor, C.J.: A
  minimum description length approach to statistical shape modeling. IEEE
  Transactions on Medical Imaging  \textbf{21}(5),  525--537 (May 2002).
  \doi{10.1109/TMI.2002.1009388}

\bibitem{eley2014black}
Eley, K.A., Watt-Smith, S.R., Sheerin, F., Golding, S.J.: {B}lack {B}one
  {M}{R}{I}: a potential alternative to ct with three-dimensional
  reconstruction of the craniofacial skeleton in the diagnosis of
  craniosynostosis. European Radiology  \textbf{24}(10),  2417--2426 (2014)

\bibitem{cates2013afib}
Gardner, G., Morris, A., Higuchi, K., MacLeod, R., Cates, J.: A
  point-correspondence approach to describing the distribution of image
  features on anatomical surfaces, with application to atrial fibrillation. In:
  2013 IEEE 10th International Symposium on Biomedical Imaging. pp. 226--229
  (April 2013). \doi{10.1109/ISBI.2013.6556453}

\bibitem{RTW:Gre91}
Grenander, U., Chow, Y., Keenan, D.M.: Hands: {A} Pattern Theoretic Study of
  Biological Shapes. Springer, New York (1991)

\bibitem{hack2010prevalence}
Hack, K., Di~Primio, G., Rakhra, K., Beaul{\'e}, P.E.: Prevalence of cam-type
  femoroacetabular impingement morphology in asymptomatic volunteers. JBJS
  \textbf{92}(14),  2436--2444 (2010)

\bibitem{hanley1982meaning}
Hanley, J.A., McNeil, B.J.: The meaning and use of the area under a receiver
  operating characteristic (roc) curve. Radiology  \textbf{143}(1),  29--36
  (1982)

\bibitem{harris2013cam}
Harris, M.D., Datar, M., Whitaker, R.T., Jurrus, E.R., Peters, C.L., Anderson,
  A.E.: Statistical shape modeling of cam femoroacetabular impingement. Journal
  of Orthopaedic Research  \textbf{31}(10),  1620--1626 (2013).
  \doi{10.1002/jor.22389}, \url{http://dx.doi.org/10.1002/jor.22389}

\bibitem{kellogg2012interfrontal}
Kellogg, R., Allori, A.C., Rogers, G.F., Marcus, J.R.: Interfrontal angle for
  characterization of trigonocephaly: part 1: development and validation of a
  tool for diagnosis of metopic synostosis. Journal of Craniofacial Surgery
  \textbf{23}(3),  799--804 (2012)

\bibitem{kokoska2000crc}
Kokoska, S., Zwillinger, D.: CRC standard probability and statistics tables and
  formulae. Crc Press (2000)

\bibitem{leys2018detecting}
Leys, C., Klein, O., Dominicy, Y., Ley, C.: Detecting multivariate outliers:
  Use a robust variant of the mahalanobis distance. Journal of Experimental
  Social Psychology  \textbf{74},  150--156 (2018)

\bibitem{mendoza2014personalized}
Mendoza, C.S., Safdar, N., Okada, K., Myers, E., Rogers, G.F., Linguraru, M.G.:
  Personalized assessment of craniosynostosis via statistical shape modeling.
  Medical Image Analysis  \textbf{18}(4),  635--646 (2014)

\bibitem{muthen1998statistical}
Muth{\'e}n, L.K., Muth{\'e}n, B.O.: Statistical analysis with latent variables.
  Mplus User’s guide  \textbf{2012} (1998)

\bibitem{oguz2016entropy}
Oguz, I., Cates, J., Datar, M., Paniagua, B., Fletcher, T., Vachet, C., Styner,
  M., Whitaker, R.: Entropy-based particle correspondence for shape
  populations. International journal of computer assisted radiology and surgery
   \textbf{11}(7),  1221--1232 (2016)

\bibitem{schlegl2017unsupervised}
Schlegl, T., Seeb{\"o}ck, P., Waldstein, S.M., Schmidt-Erfurth, U., Langs, G.:
  Unsupervised anomaly detection with generative adversarial networks to guide
  marker discovery. In: International conference on information processing in
  medical imaging. pp. 146--157. Springer (2017)

\bibitem{sica2006bias}
Sica, G.T.: Bias in research studies. Radiology  \textbf{238}(3),  780--789
  (2006)

\bibitem{styner2006spharm}
Styner, M., Oguz, I., Xu, S., Brechbuehler, C., Pantazis, D., Levitt, J.,
  Shenton, M., Gerig, G.: Framework for the statistical shape analysis of brain
  structures using spharm-pdm  (07 2006)

\bibitem{thompson1942growth}
Thompson, D.W., et~al.: On growth and form. On growth and form.  (1942)

\bibitem{uebersax1993latent}
Uebersax, J.S., Grove, W.M.: A latent trait finite mixture model for the
  analysis of rating agreement. Biometrics pp. 823--835 (1993)

\bibitem{weinzweig2003metopic}
Weinzweig, J., Kirschner, R.E., Farley, A., Reiss, P., Hunter, J., Whitaker,
  L.A., Bartlett, S.P.: Metopic synostosis: Defining the temporal sequence of
  normal suture fusion and differentiating it from synostosis on the basis of
  computed tomography images. Plastic and Reconstructive Surgery
  \textbf{112}(5),  1211--1218 (2003)

\bibitem{wood2016name}
Wood, B.C., Mendoza, C.S., Oh, A.K., Myers, E., Safdar, N., Linguraru, M.G.,
  Rogers, G.F.: What’s in a name? accurately diagnosing metopic
  craniosynostosis using a computational approach. Plastic and Reconstructive
  Surgery  \textbf{137}(1),  205--213 (2016)

\bibitem{yee2015classification}
Yee, S.T., Fearon, J.A., Gosain, A.K., Timbang, M.R., Papay, F.A., Doumit, G.:
  Classification and management of metopic craniosynostosis. Journal of
  Craniofacial Surgery  \textbf{26}(6),  1812--1817 (2015)

\bibitem{zimmerer2019unsupervised}
Zimmerer, D., Isensee, F., Petersen, J., Kohl, S., Maier-Hein, K.: Unsupervised
  anomaly localization using variational auto-encoders. In: International
  Conference on Medical Image Computing and Computer-Assisted Intervention. pp.
  289--297. Springer (2019)

\end{thebibliography}

\end{document}